\documentclass[sigconf]{acmart}
\AtBeginDocument{%
  }

\usepackage{xcolor}
\usepackage{booktabs}
\usepackage{array}
\usepackage{multirow}
\usepackage{afterpage}
\usepackage{subcaption}
\usepackage{booktabs}
\usepackage{siunitx}

\copyrightyear{2026}
\acmYear{2026}
\setcopyright{none}
\setcctype{by}
\acmConference[LAK 2026]{LAK26: 16th International Learning Analytics and Knowledge Conference}{April 27-May 01, 2026}{Bergen, Norway}
\acmBooktitle{LAK26: 16th International Learning Analytics and Knowledge Conference (LAK 2026), April 27-May 01, 2026, Bergen, Norway}
\acmPrice{}
\acmDOI{10.1145/3785022.3785029}
\acmISBN{979-8-4007-2066-6/2026/04}

\begin{document}

\title{The StudyChat Dataset: Analyzing Student Dialogues With ChatGPT in an Artificial Intelligence Course}

\author{Hunter McNichols}
\email{wmcnichols@umass.edu}
\affiliation{%
  \institution{University of Massachusetts Amherst}
  \city{Amherst}
  \country{United States}
}

\author{Fareya Ikram}
\email{fikram@umass.edu}
\affiliation{%
  \institution{University of Massachusetts Amherst}
  \city{Amherst}
  \country{United States}
}

\author{Andrew Lan}
\email{andrewlan@cs.umass.edu}
\affiliation{%
  \institution{University of Massachusetts Amherst}
  \city{Amherst}
  \country{United States}
}

\renewcommand{\shortauthors}{McNichols et al.}

\begin{abstract}
The widespread availability of large language models (LLMs), such as ChatGPT, has significantly impacted education, raising both opportunities and challenges. Students can frequently interact with LLM-powered, interactive learning tools, but their usage patterns need to be observed and understood. We introduce StudyChat, a publicly available dataset capturing real-world student interactions with an LLM-powered tutoring chatbot in a semester-long, university-level artificial intelligence (AI) course. We deploy a web application that replicates ChatGPT's core functionalities, and use it to log student interactions with the LLM while working on programming assignments. We collect 16,851 interactions, which we annotate using a dialogue act labeling schema inspired by observed interaction patterns and prior research. We analyze these interactions, highlight usage trends, and analyze how specific student behavior correlates with their course outcome. We find that students who prompt LLMs for conceptual understanding and coding help tend to perform better on assignments and exams. Moreover, students who use LLMs to write reports and circumvent assignment learning objectives have lower outcomes on exams than others. StudyChat serves as a shared resource to facilitate further research on the evolving role of LLMs in education \footnote{https://huggingface.co/datasets/wmcnicho/StudyChat}.
\end{abstract}



%     \centering
%     \includegraphics[width=0.99\linewidth]{figs/studychat_overview_wide.png}
%     \caption{An overview of the dataset collection and analysis pipeline present in our work.}
%     % An overview of our study. Student–LLM conversations are collected during programming assignments and labeled by a dialogue act schema. These conversations are used to form student behavior features which are analyzed to investigate correlations between interaction patterns and learning outcomes.}
%     \Description{An overview of the dataset collection and analysis pipeline present in this work.}
%     \label{fig:placeholder}

\maketitle

\section{Introduction}
Advances in large language models (LLMs) have had major ramifications on almost all levels of education. Many LLM-powered educational technologies have recently been developed, leveraging the generative capabilities of these models to create new course content \cite{macneil2022generating}, assignments \cite{MacNeil_2023}, and interactive learning experiences \cite{jin2024teach,taneja2024jill}. These interactive technologies enable rich engagement with course material by allowing students to engage in on-demand conversations with an interactive tutor and receive personalized feedback on their work. However, while the potential use cases for LLMs in education are vast, many educators are understandably concerned about student misuse of LLM chatbot tools, such as ChatGPT \cite{lau2023ban}. While a motivated learner can use these tools positively, an unmotivated student could use them to complete their assignments, circumventing the learning objectives of the assignment, or become over-reliant on the model \cite{bastani2024generative}. 

% \begin{figure}[t]
%     \centering
%     \includegraphics[width=0.99\linewidth]{figs/studychat_overview_wide.png}
%     \caption{Overview of our study. Student–LLM conversations are collected during programming assignments and labeled by dialogue acts. These conversations are used to form student behavior features which are analyzed and clustered to investigate connections between interaction patterns and learning outcomes.}
%     \label{fig:placeholder}
% \end{figure}

This dichotomy between learning opportunity and academic risk necessitates an exploration into understanding student behavior with LLM chatbot tools. A more nuanced understanding of this behavior could inform further development of LLM-based Intelligent Tutoring Systems (ITS) that dynamically adapt to students, encouraging learning-focused behaviors and discouraging counterproductive ones. Furthermore, a deeper understanding of student behavior can lead to more robust student models, which better estimate knowledge states, misconceptions, and learning trajectories. For example, such a model could potentially detect when a student is overly reliant on LLM responses and flag this behavior to instructors in real-time, or be used to help develop assessments that are specifically designed to test for this over-reliance.

\begin{figure*}[ht]
    \centering
    \includegraphics[width=0.85\linewidth]{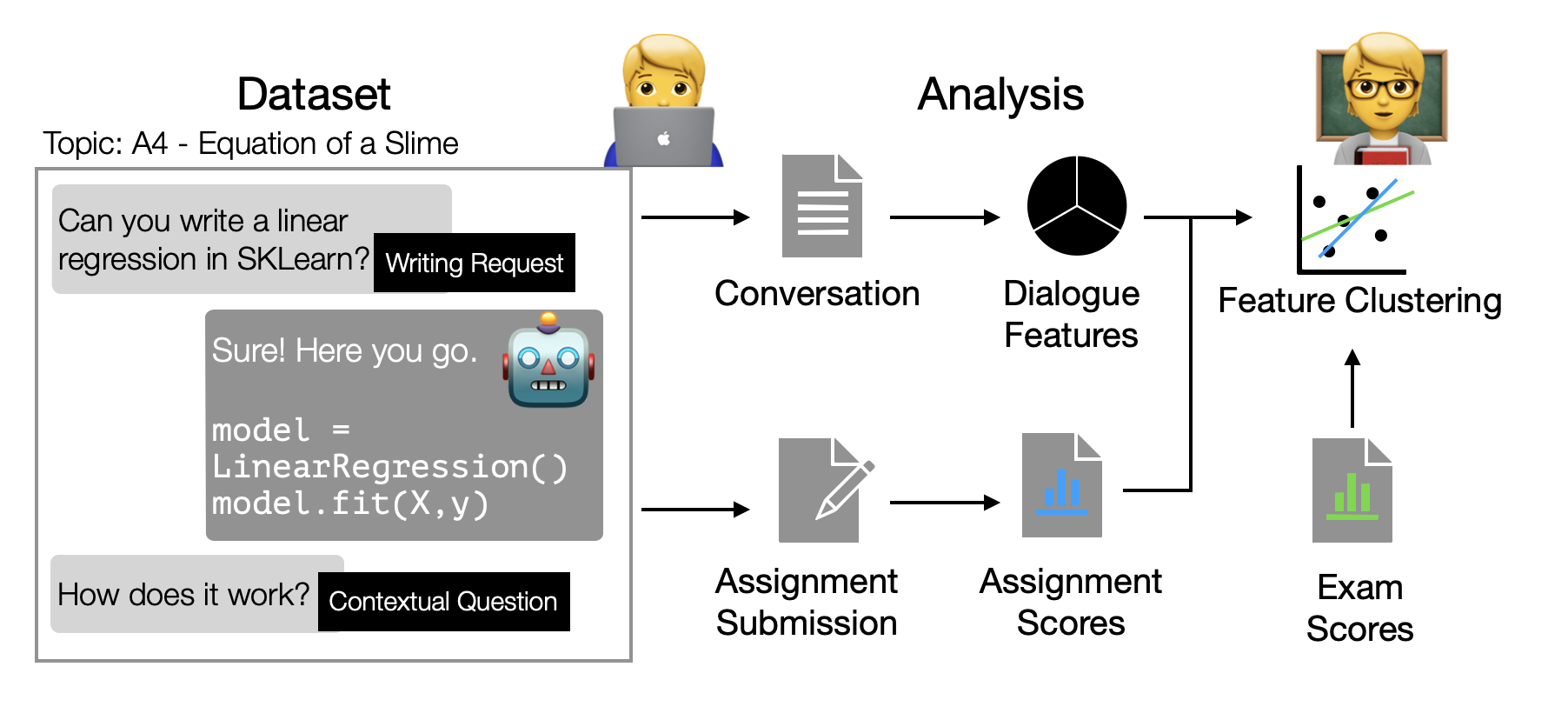}
    \caption{An overview of the dataset collection and analysis pipeline present in our work.}
    % An overview of our study. Student–LLM conversations are collected during programming assignments and labeled by a dialogue act schema. These conversations are used to form student behavior features which are analyzed to investigate correlations between interaction patterns and learning outcomes.}
    \Description{An overview of the dataset collection and analysis pipeline present in this work.}
    \label{fig:placeholder}
\end{figure*}

In this work, we conduct a year-long case study to explore student behavior with an LLM, specifically in the context of upper-division, university-level Computer Science (CS) education. We develop a web application with the same core features and underlying LLM as ChatGPT, and deploy it in an introductory-level course on Artificial Intelligence (AI). We encourage students to use this tool, without restrictions, on programming assignments over the course of the semester. Across two semesters and with student consent, we recorded 2,214 student-LLM conversations from 203 students across 7 unique assignments, totaling 16,851 unique utterances. We also collect 924 programming assignment submissions from 158 students who further agreed to share their work. In addition, we annotate these conversation utterances with a dialogue act (DA) labeling schema; the schema is inspired by existing studies on co-programming behavior and our systematic review of the conversations. We leverage an LLM prompting-based approach to do this annotation at scale and validate this approach with a human evaluation.  

We explore how students engage with an LLM during programming assignments. In doing so our primary research objective is to uncover patterns of behavior across students and analyze if these trends correlate with course outcomes. To do this, we analyze these DA labels by performing regression and clustering analysis. Our key findings are: 1) students primarily use the LLM to ask questions, but our findings suggest that some question types are negatively correlated with outcome, suggesting student confusion possibly due to LLM hallucination; 2) certain behavior patterns such as asking conceptual questions or help in code writing are positively associated with both assignment and held out exam scores;  3) students with low and high usage of LLMs have similar average course outcome but students that use that LLMs with high frequency tend to have less variance in a more narrow positive range; 4) clustering analysis reveals students who use the LLM primarily for code writing and not other writing requests tend to score higher on both assignments and exams. These findings provide data-driven insights into patterns of student LLM behaviors in a university-level CS course. We release this dataset to the broader research community to foster collaboration and exploration into LLMs' impact on AI and CS education.

\section{Related Works}

\subsection{ChatGPT Classroom Studies}
Since its release in November 2022, there has been a vast amount of research interest into how ChatGPT can be utilized in classroom settings \cite{zheng2025studentsrelyaianalysis, ammari2025students}. Studies have explored various applications of ChatGPT for course exercises \cite{xiao2023evaluating, kumar2023improving}, code explanations \cite{chen2023gptutor}, multiple-choice distractors \cite{feng2024exploring, parikh-etal-2025-lookalike}, at-risk student detection \cite{liu2024feasibility}, and more \cite{khan2023assessing,li2024paper,hsieh2024enhancing,dunder2024kattis,helden2025got,moore2022assessing,lee-etal-2024-exploring-automated, lee2025text, duan2025automated}. Direct classroom studies have also explored how ChatGPT-powered applications can answer student questions as a virtual teaching assistant \cite{taneja2024jill}, help students learn by teaching chatbots to code \cite{jin2024teach}, and even provide coaching and feedback to instructors on content delivery \cite{wang2023chatgpt}.

Closely related to this work are a handful of case studies exploring student use of ChatGPT in university-level CS classrooms. Ghimire et. al explore student use of ChatGPT for two assignments in a CS1 programming course, finding students often ask questions of the system instead of directly seeking solutions \cite{ghimire2024coding}. Ma et. al also perform a case study where CS1 students were allowed to use ChatGPT freely on in-class exercises and on a course project, but only for conceptual questions on homework \cite{ma2024enhancing}. Additionally, Brender et. al perform a case study on how graduate students in a robotics course utilize ChatGPT on a voluntary assignment, controlling for learning outcomes and clustering behavior into broad categories \cite{brender2024s}. Analysis of student prompts in these studies have noted a student trend to use models to ask questions more than directly writing code solutions, which is consistent with our observations and those by \cite{bhalerao2024my}. Surveys on student perceptions towards LLMs have also been conducted, which suggest that CS student adoption of LLMs is nuanced and not universal \cite{weber2024measuring}. Similarly, we observe that some students use the system heavily while most students engage moderately with the model, revealing a range of adoption behaviors.
% \cite{zheng2025studentsrelyaianalysis, ammari2025students, fu2025mining}

\subsection{Education Discourse Analysis}
It has been demonstrated by many that tutoring is a highly effective form of instruction \cite{cohen1982educational}. Thus, human tutoring dialogue has been studied extensively \cite{graesser1995collaborative}. In recent years, DA classification on conversation turns is often used in this discourse analysis \cite{chen-etal-2011-exploring, ezen2015classifying}. DA labels classify a participant's utterance by the function it serves in the conversation. When DA labeling is applied to an entire corpus of dialogues, it enables discourse-level analysis of conversation patterns that are frequent or significant. In education, significant patterns are often those associated with better learning outcomes, as such existing works have studied such patterns in tutor-student or student-student learning environments \cite{graesser1995collaborative,earle2023confusion,ikram2025exploring}. Existing works also analyze human-machine classroom conversations between students and ITS chatbots \cite{vail2014identifying,graesser2004autotutor,fisher2020livehint}. These works provide promising insights which can inform more effective tutoring pedagogy \cite{rus2017analysis} and interactive learning tools \cite{chen-etal-2011-exploring}.

While DA analysis is an attractive approach to discourse analysis, the labeling process poses many challenges, solutions to which constitute a large body of work \cite{boyer2011affect,tan2023does,ezen2015understanding,lin2023robust,boyer2010dialogue,kumaran2023improving}. First, there exists the task of annotating a large corpus of conversations. To automate such a task, various machine learning based approaches to this have been used, such as Hidden Markov Models \cite{boyer2010dialogue}, decision trees \cite{samei2014context}, and Markov Random Fields \cite{ezen2015understanding}. In our work, we employ a LLM prompt-based approach to DA labeling, similar to that of \cite{wang2023can}. Second, there exists the challenge of robust schema design, where existing work has looked into refining prior labeling schemata to be less ambiguous \cite{vail2014identifying} and to try to develop a classroom agnostic labeling schema \cite{hennessy2016developing}.

\subsection{Education Dialogue Datasets}
In an effort to collectively improve data-driven learning technologies, there have been recent efforts to expand the amount of education dialogue datasets in a variety of learning domains. Datasets such as TalkMoves and NCTE directly capture and annotate classroom discourse in K-12 lessons \cite{suresh2022talkmoves,demszky2022ncte}. Other works employ crowd workers or leverage LLMs to simulate the student or teacher role, such as MathDial \cite{macina2023mathdial}, Teacher-Student Chatroom Corpus (TSCC) \cite{caines2020teacher}, Conversational Instruction with Multi-responses and Actions (CIMA) \cite{stasaski2020cima}, and GrounDialog \cite{zhang2023groundialog}. Our work contributes to this growing body of research by providing a large-scale, annotated dialogue dataset in a university-level CS course, offering insights into how students engage with LLM-chatbot tools in a real learning environment.

\section{Dataset}

In this section, we detail the process by which the StudyChat dataset was formed and the classroom context.

\subsection{Participants and Course Context}
Our IRB-approved study was conducted among students aged 18 or above, who were enrolled in an AI course at the University of Massachusetts Amherst in the Fall 2024 and Spring 2025 semesters. Across both semesters, 295 students were enrolled in the course, of which 203 agreed to participate in the study. Students who did not participate were also given equivalent access to use the tool over the semester, but their data was not collected. Students were in their 3rd/4th years, had completed their foundational Computer Science course sequences and were taking higher-level elective courses. 

Each week, students attend two 75-minute lectures and independently complete assignments related to the topics covered in lecture. The course consists of 3 modules taught over 15 weeks. The first module covers the foundations of the rational agent approach (focused on applied search algorithms), the second module covers fundamentals of machine learning with a focus on deep learning and neural networks, and the final module covers recent topics in applied AI. During each module, students complete 2-3 Python programming assignments related to the lecture material. In addition to the assignments, students were assessed via 2 end-of-module exams, and a final exam. 

The programming assignments are Python-based and can be broadly categorized into two types. The first assignment type is a tutorial, where students are given incomplete Jupyter notebooks that guide students to learn fundamental operations in applied machine learning Python libraries, including Pandas, SciKit Learn, and PyTorch. The second assignment type is an applied scenario, where students are given Python starter code that they extend by applying an algorithm or concept taught in class. For example, in assignment 2, students are tasked with creating a text autocomplete engine by implementing various search algorithms to traverse graphs to find word completions. Assignments were altered slightly between semesters, but generally cover the same topics. All assignment instruction details, starter code, and 924 graded submissions are included in the dataset. 

\begin{figure}[t]
    \centering
    \includegraphics[width=0.99\linewidth]{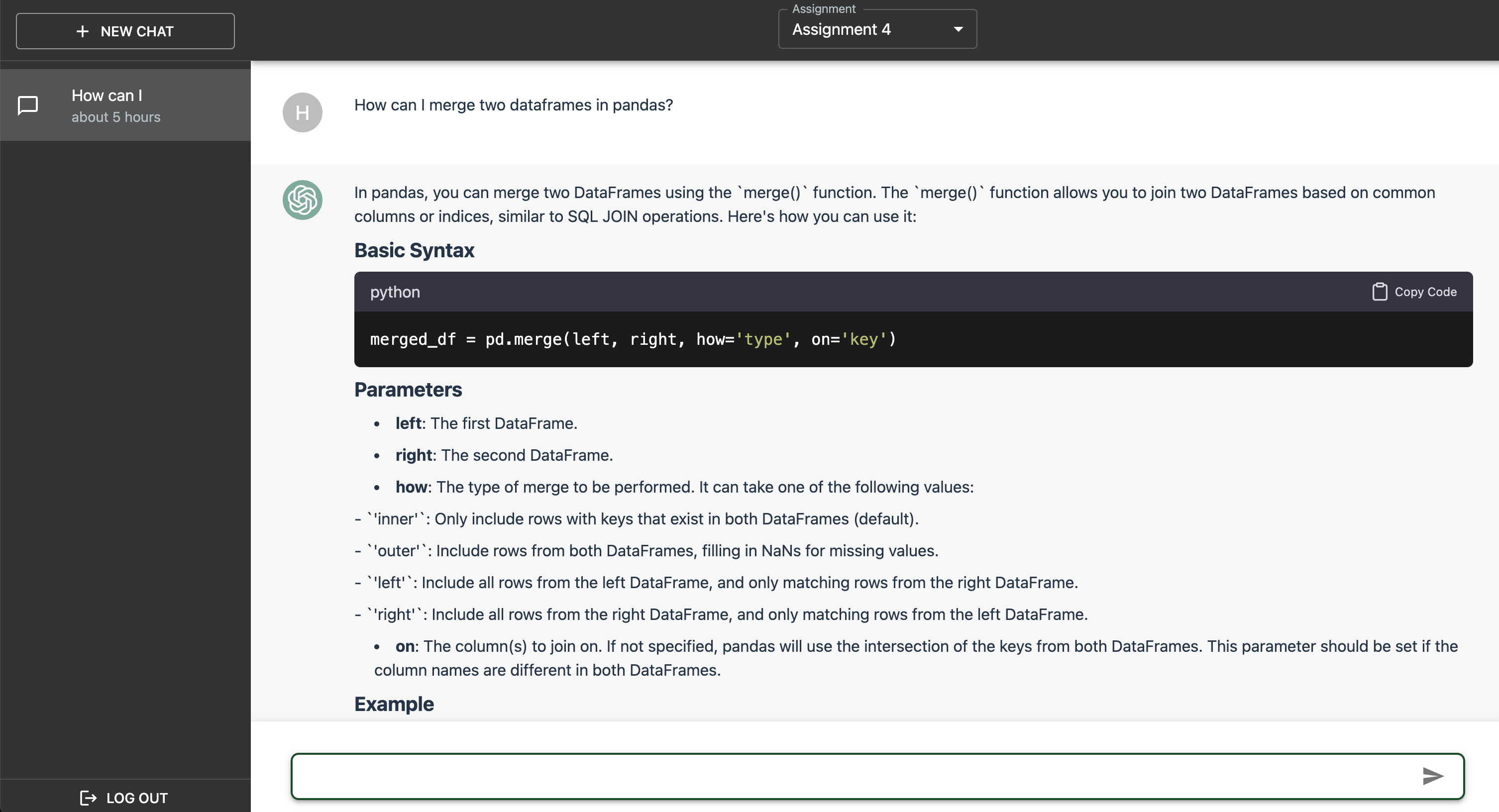}
    \caption{The web application used by students in this study. The user interface mirrors the functionality of ChatGPT.}
    \label{fig:383gpt_interface}
\end{figure}

\subsection{Data Collection}

In order to effectively collect student utterances, we developed and deployed a web application designed to function similarly to consumer LLM chatbot interfaces, shown in Figure~\ref{fig:383gpt_interface}. As our backing chatbot LLM, we used \texttt{gpt-4o-mini}, the LLM for the free tier of ChatGPT at the time of this study \cite{gpt4o-mini}. We evaluated this model by testing its capabilities with the libraries and concepts covered in the assignments, and found that the model generally provides accurate guidance on the assignments and implementations that are often correct. Students were instructed that they could use this ChatGPT-based tool, without restrictions, on all assignments in the course. It was emphasized that their approach to using the tool would not affect their grade, and they were encouraged to use the tool in any manner of their choosing. We used the default "You are a helpful assistant" system prompt, which does not discourage answer-seeking behavior, and did not provide the chatbot with any course materials for reference. Our goal was to encourage students to use our platform instead of seeking an outside application for assignment help, and that the collected interactions would more accurately reflect how students generally use non-instructional, consumer chatbots. 

% Define a custom column type for ragged right alignment inside p{} columns
\newcolumntype{L}[1]{>{\raggedright\arraybackslash}p{#1}}
\begin{table*}[ht!!]
    \centering
    \renewcommand{\arraystretch}{1.2}
    \setlength{\tabcolsep}{6pt}
    \caption{Dialogue act schema for categorizing student–LLM interactions, with counts by semester.}
    \label{tab:dialog_act_schema_counts}
    \begin{tabular}{L{0.20\linewidth} L{0.40\linewidth} r r r}
        \toprule
        \textbf{Broad Category} & \textbf{Specific Dialogue Acts} & \textbf{f24} & \textbf{s25} & \textbf{Total} \\
        \midrule
        Writing & Write Code, Write English, Conversion, Summarize, Other 
                & 1664 & 2672 & 4336 \\
        \midrule
        Editing & Edit Code, Edit English, Other 
                & 153 & 265 & 418 \\
        \midrule
        Contextual Questions & Assignment Clarification, Code Explanation, Interpret Output, Other 
                & 1216 & 1722 & 2938 \\
        \midrule
        Conceptual Questions & Programming Language, Python Library, Computer Science, Programming Tools, Mathematics, Other Concept 
                & 2204 & 2995 & 5199 \\
        \midrule
        Verification & Verify Code, Verify Report, Verify Output, Other 
                & 293 & 354 & 647 \\
        \midrule
        Context & Assignment Information, Error Message, Code, Other 
                & 1191 & 1837 & 3028 \\
        \midrule
        Off Topic & Chit-Chat, Greeting, Gratitude, Other 
                & 128 & 122 & 250 \\
        \midrule
        Misc & Other 
                & 15 & 20 & 35 \\
        \bottomrule
    \end{tabular}
\end{table*}
\subsection{Filtering of Student Identifiable Information}
To remove any Personally Identifiable Information (PII) from the conversations and assignment submissions, we systematically analyze the conversations to uncover patterns of PII exposure. We first randomly sample 12 conversations from each course topic and read through the conversations, scanning for traces of PII. Our main finding of potentially sensitive student information is in the form of pasted terminal outputs and error messages, which sometimes contain PII, such as a user's directory or GitHub account name. Compiling these common patterns, we then develop a regular expression-based script to remove these patterns and any direct matches to known user PII in the course roster. Across the entire dataset, this process removes 6,413 cases of possible PII exposure. Then, we sample 12 additional conversations from each topic and performed the same procedure. In our second pass of the second set of conversations, we did not find any additional patterns or cases of PII exposure. We acknowledge that, although seemingly effective, this process does not guarantee the removal of all PII exposures.

% \caption{A boxplot showing the distribution of interactions by assignment across each semester (left) and distribution of dialogue act labels (right).}
% \label{fig:combined_human_exp}
% \end{figure}

\subsection{Data Statistics}
Across all conversations, we collect 16,851 student utterances and corresponding LLM responses over 7 graded assignments. The mean conversation length in the dataset is 7.6 utterances, with a median of 4.0 utterances per conversation. On average, each student has 83 utterances and 10.9 distinct conversations with the chatbot. 

\subsection{Dialogue Act Annotation}
In order to better understand common behavior and discourse-level trends in the conversations, we develop a DA labeling schema for student turns in the conversations. The schema is two-leveled, with 8 broad labels and 31 specific labels, and is developed on top of existing work \cite{wood2018detecting}, which conducted a ``Wizard of Oz'' experiment with programmers who were debugging alongside a simulated virtual assistant. It also mirrors our own observations after systematically reviewing the conversations and aligns with those of Ma. et al \cite{ma2024enhancing}. The schema distinguishes the various ways students use an LLM to assist in writing aspects of their assignments and ask questions. We report total counts of each broad label in Table~\ref{tab:dialog_act_schema_counts}.
% total counts of specific labels in Table~\ref{tab:dialogue_act_distribution}, and detail full descriptions of the schema in Table~\ref{tab:dialog_act_schema}.

To validate the accuracy of this labeling, we first sample a set of 150 utterances and develop an initial schema with labeling guidelines, including examples and descriptions for each DA. We then provide them to a research assistant, who independently labels the same set given our initial schema. We iterate on our labeling instructions and schema by computing the agreement between the two annotators and identifying sources of disagreement. After this iteration, our inter-rater agreement reached a Cohen's kappa \cite{cohen1960coefficient} of 0.910 for broad level labels and 0.788 for specific labels, indicating almost perfect agreement on the broad level of DA and substantial agreement on the specific level \cite{landis1977measurement}. To validate this labeling, we provided these final instructions to two additional annotators who independently label the same set of utterances. We observed an average 4-way inter-rater agreement Cohen's kappa of 0.740 for broad level agreement and  0.575 for specific level agreement, indicating substantial agreement on broad level labeling and moderate agreement on specific labeling.

To apply this labeling at scale to the remainder of the dataset and since manual labeling is extraneous, we perform an LLM-prompting-based approach similar to prior work \cite{wang2023can, yin2025scaling}. We prompt the LLM with detailed instructions on the task, the annotation instructions, assignment instructions, and the entire conversation up until the student turn to be labeled, and ask the LLM to provide the label based on our schema, along with a justification. We utilize \texttt{GPT4.1} to perform this labeling, which we applied to the entire PII-filtered dataset.

To validate the LLM-generated labels, we again systematically sample 154 utterances, distinct from those used in developing the initial labeling schema. We provide them to the same 4 human annotators without any further discussion or instructions. The resulting human inter-annotator agreement on the held out set is 0.578 on broad labels and 0.538 on specific labels, indicating moderate agreement between human annotators on the held out set. The average human-LLM agreement on the held-out set is 0.5819 on the broad labels and 0.4927 on the specific labels, indicating moderate agreement on par with human inter-annotator agreement. This drop in agreement is somewhat expected, since the new utterances in the held-out set present new ambiguous, overlapping utterances to which multiple labels may apply.

% \begin{table}[t]
% \centering
% \renewcommand{\arraystretch}{1.1}
% \setlength{\tabcolsep}{6pt}
% \caption{Total counts per broad dialogue act label in StudyChat. Further breakdown of specific label counts can be found in Table~\ref{tab:dialogue_act_distribution}.}
% \label{tab:dialogue_act_totals}
% \begin{tabular}{l r r r}
% \toprule
% \textbf{Label} & \textbf{f24} & \textbf{s25} & \textbf{Total} \\
% \midrule
% Conceptual Questions & 2204 & 2995 & 5199 \\
% Writing  Request     & 1664 & 2672 & 4336 \\
% Provide Context      & 1191 & 1837 & 3028 \\
% Contextual Questions & 1216 & 1722 & 2938 \\
% Verification         &  293 &  354 &  647 \\
% Editing              &  153 &  265 &  418 \\
% Off Topic            &  128 &  122 &  250 \\
% Misc                 &   15 &   20 &   35 \\
% \bottomrule
% \end{tabular}
% \end{table}

\begin{table*}[t]
\centering
\renewcommand{\arraystretch}{1.15}
\setlength{\tabcolsep}{5pt} % slightly narrower spacing
\caption{Statistically significant regression coefficients ($p < 0.05$) for predicting assessment outcomes from interaction features, grouped by semester and feature set. Only significant results are listed, and targets (assignments and exams) abbreviated.}
\label{tab:regression_summary}
\begin{tabular}{lll lrrr}
\toprule
\textbf{Semester} & \textbf{Feature Set} & \textbf{Feature Name} & \textbf{Target} & \textbf{Coef.} & \textbf{$p$} & \textbf{$n$} \\
\midrule
\multirow{8}{*}{F24}
    & Broad & \multicolumn{5}{l}{No significant coefficients} \\
    \cmidrule(l){2-7}
    & \multirow{7}{*}{Specific}
        & Contextual Questions: Code Explanation      & A3 & $-0.0065$ & $0.021$ & $72$ \\
    &   & Contextual Questions: Other                & A3 & $-0.0156$ & $0.034$ & $72$ \\
    &   & Off Topic: Chit-Chat                & A3 & $0.0235$  & $0.043$ & $72$ \\
    &   & Provide Context: Code               & A4 & $-0.0426$ & $0.001$ & $66$ \\
    &   & Verification: Other                 & A4 & $-0.0291$ & $0.007$ & $66$ \\
    &   & Conceptual Questions: Other Concept        & A5 & $-0.0575$ & $0.025$ & $64$ \\
    &   & Writing Req: Other                  & A5 & $-0.0590$ & $0.033$ & $64$ \\
\midrule
\multirow{15}{*}{S25}
    & \multirow{3}{*}{Broad}
        & Conceptual Questions                        & E1 & $0.0133$  & $0.024$ & $104$ \\
    &   & Editing Req                          & E1 & $0.0296$  & $0.017$ & $104$ \\
    &   & Contextual Questions                        & A5 & $-0.0020$ & $0.005$ & $72$ \\
    \cmidrule(l){2-7}
    & \multirow{12}{*}{Specific}
        & Conceptual Questions: Python Library        & E1 & $0.0234$  & $0.010$ & $104$ \\
    &   & Contextual Questions: Assignment Clarif.    & A3 & $-0.0034$ & $0.021$ & $103$ \\
    &   & Conceptual Questions: Other Concept        & A3 & $0.0095$  & $0.029$ & $103$ \\
    &   & Writing Req: Conversion              & E2 & $0.0028$  & $0.045$ & $119$ \\
    &   & Contextual Questions: Assignment Clarif.    & A5 & $-0.0072$ & $0.039$ & $72$ \\
    &   & Contextual Questions: Code Explanation      & A5 & $0.0009$  & $0.044$ & $72$ \\
    &   & Contextual Questions: Other                 & A5 & $-0.0075$ & $0.017$ & $72$ \\
    &   & Conceptual Questions: Other Concept        & A5 & $0.0016$  & $0.019$ & $72$ \\
    &   & Conceptual Questions: Mathematics           & A6 & $-0.1062$ & $0.002$ & $74$ \\
    &   & Off Topic: Gratitude                 & E3 & $-0.0337$ & $0.025$ & $119$ \\
\bottomrule
\end{tabular}
\end{table*}

\section{Results and Analysis}
In this section, we study the relationship between student interactions with the LLM chatbot and their course outcomes.

\subsection{Dialogue Acts as Assessment Predictors}

We postulate that student interactions with LLMs during assignments will have an observable influence on student course outcomes. To study this hypothesis, we run a series of regression models and measure the varying effects of adding LLM usage features of increasing specificity. First, we use a single feature of the total count of utterances, then a series of 8 features corresponding to the broad DA label counts, and finally a series of 31 features corresponding to the specific DA label counts. For all models, we use a dependent variable of the score of an assessment normalized to the range $[0, 1]$. In this analysis, we define assessment as one of the 7 assignments (a1-a7) or 3 course exams (e1-e3). For our baseline model, we use one independent variable, the running average of prior assessment scores in the range $[0,1]$, a well-studied predictor of post assessment score \cite{hallberg2018pretest}. We run each model 18 times, for 9 assessments across both semesters, excluding a1, which has no prior assessment. For the baseline model, the mean $R^2$ values for all regressions are $0.17343$ and $0.17281$, for the fall and spring semesters, respectively.

For our first model, we add one additional independent variable to the linear regression model: the total count of all utterances, inclusive of the current assignment. For exams, we sum the utterances for all assignments that were due before the exam. For these regressions, the mean $R^2$ was $0.1847$ for the fall and $0.1860$ for the spring, with mean $R^2$ improvements of $0.0064$ and $0.0178$, respectively. We run F-tests for each improvement, and we find that for one assessment, spring semester a5, the addition of the usage count independent variable leads to a statistically significant improvement in explained variance, with an increase of $0.0242$, $F=5.82$, $p=0.02$. In the other cases, we found no statistically significant improvement. 

For our next two models, we explore the influence of incorporating DA information into the regression. We expand our independent variables to include features based on the two levels of DA behavior labels, using 8 broad-labeled DAs or 31 specific-labeled DAs, respectively. Specifically, we augment the baseline regression by including the counts of utterances for each of the DA labels present in the assessment target. In the same manner as above, for exams, we aggregate all assignment-specific DA counts that were due prior to the exam. With the inclusion of broad DA label counts, we observe that this inclusion has a mean $R^2$ of $0.2529$ for the fall and $0.2288$ for the spring semester, with a mean improvement of $0.0746$ and $0.0606$, respectively. These $R^2$ values indicate that DA features contribute explanatory power, but F-Tests conducted for each regression do not indicate results that are statistically significant, with the highest $R^2$ improvement of $0.19$ and resulting F-Score of $F=1.81, p=0.09$. Repeating the same experiment with the specific DA label count features, we observe that this inclusion increases the mean $R^2$ to $0.4868$ for the fall semester and $0.4012$ for the spring semester, with a mean  $R^2$ increase of $0.3086$ and $0.2330$, respectively. Similar to the broad features, these $R^2$ values indicate that DA features contribute explanatory power, but F-Tests conducted for each regression do not indicate results that are statistically significant. One assessment, a4 from the fall semester, has a borderline significant improvement of $0.4$ with an F-Score of $F=1.77, p=0.05$.

Our findings in these regression analyses have mixed results in support of our hypothesis. The analysis indicates that the explanatory power of a linear regression model increases significantly when moving from a simple count of utterances to a more nuanced, qualitative representation based on DAs. However, the relative explanatory power compared to a simple, prior performance-based explainer of student outcome is only statistically significant when the utterances are represented as aggregate counts, which may simply be a confounder for student engagement or participation. This discrepancy suggests that the DA-based models are likely overfitting the data or are statistically underpowered by a limited sample size. While the relationship between dialogue patterns and outcome likely exists, a more sophisticated feature selection method and modeling approach would be required to build a model that is highly predictive and robust in explaining student assessment course outcomes. 

We further analyze the more granular relationships between DA and assessment outcomes by examining significant coefficients in our regression models. Specifically, for the broad and specific DA regression models, we run F-Tests to compute the significance of the coefficients for each target. Across the 36 regressions, 20 coefficients were found to be statistically significant, which we report in Table~\ref{tab:regression_summary}. For the fall semester, we observe significant coefficients only from specific-level DA regressions and only for assignment targets. Most coefficients are negatively correlated with assignment scores and largely fall into \textit{other} specific DA labels, with the only positive significant correlation attributed to an \textit{off-topic} DA. One possible explanation for this observation is that, compared to other DAs, these labels occur relatively infrequently in the distribution, and the linear regression was likely skewed by a small subset of student behavior patterns that were uncommon. 

% \begin{figure*}[ht]
%     \centering
%     \includegraphics[width=0.9\linewidth]{figs/side-by-side-reg_v2.png}
%     \caption{Average assignment scores (LLM-supported) vs. exam scores (No-LLM) for Fall 2024 (left, circles) and Spring 2025 (right, squares), grouped by interaction level: bottom 10\% (red), middle 80\% (blue), and top 10\% (green). Grey dashed line shows equal outcome, and red line shows the data trend for each semester; top users cluster at higher scores, and bottom users show a wider spread and more low outliers across both semesters.}
%     \label{fig:llm_nollm}
% \end{figure*}
The spring semester's significant coefficients are mostly distinct from the Fall, and seem to have potentially more meaningful pedagogical takeaways. One consistent theme across both semesters is that \textit{Contextual Questions} acts are negatively correlated with assignment outcomes. Unique to the Spring regressions are statistically significant coefficients when the target is an exam score and the inclusion of broad-level DA features. We observe the broad feature for \textit{Conceptual Questions} and \textit{Editing Request} as positively correlated with e1 outcomes, indicating a learning-focused usage of LLMs being positively correlated with assessment outcomes. Conversely, the \textit{Contextual Questions} broad level DA was negatively correlated with Assignment 5 outcome. Since the \textit{Contextual Questions} label corresponds to student questions that are related to the specifics of the current conversation, this negative correlation is perhaps an indicator of student confusion. On the other hand, the \textit{Conceptual Questions} label indicates a student asking a general knowledge question to the chatbot instead of something specific to the course. One possible takeaway from this finding is that educators can foster student use of LLMs for general knowledge and well-established concepts, but discourage asking for course-specific help. These suggestions may help to strike a balance for educators in fostering a modern technology-driven classroom, while preventing students from developing misconceptions from inaccurate chatbot responses.

In investigating the specific DAs in the spring semester, we see Assignment 5 as a consistent target and also observe specific contextual question features as having negative correlation, with the exception of \textit{Contextual Question - Code Explanation}, which has a slight positive correlation. For the spring dataset regressions, we also see different specific types of \textit{Conceptual Questions} being mostly positively correlated with scores. One notable exception is that \textit{Conceptual Question - Mathematics} is a strong negative indicator of a6 score. In this assignment, students were tasked with building an $n$-gram language model, which involved computing probability tables and deriving equations for the model. This correlation suggests that students who relied heavily on the model to understand these equations and derivations did not score well, potentially deriving incorrect solutions from the model or receiving inaccurate guidance. This observation suggests that LLM conceptual knowledge and explanations are inconsistent and that LLMs, or at least \texttt{gpt4o-mini}, are poor at explaining or deriving math equations. This finding suggests that educators should advise students to be cautious with LLM explanations and to inform them that while LLMs may have accurate general knowledge, the procedural derivations of these models should be carefully validated. 

\begin{figure*}[t]
    \centering
    \includegraphics[width=0.9\linewidth]{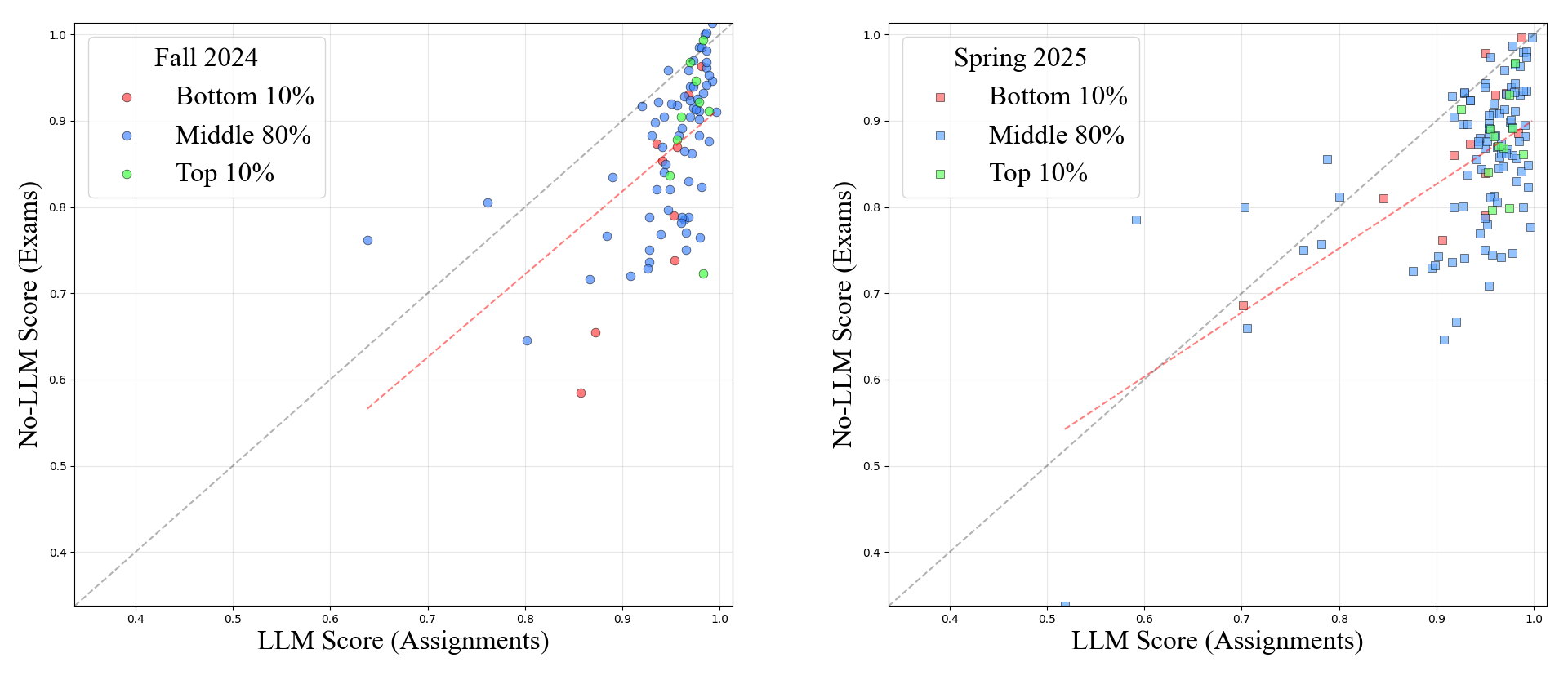}
    \caption{Average assignment scores (LLM-supported) vs. exam scores (No-LLM) for Fall 2024 (left, circles) and Spring 2025 (right, squares), grouped by interaction level: bottom 10\% (red), middle 80\% (blue), and top 10\% (green). Grey dashed line shows equal outcome, and red line shows the data trend for each semester; top users cluster at higher scores, and bottom users show a wider spread and more low outliers across both semesters.}
    \label{fig:llm_nollm}
\end{figure*}

\subsection{Low and High Usage Users}

To understand how the level of user interaction influences outcome, we examine students in the bottom 10\%, middle 80\%, and top 10\% in terms of total utterances over the course of the semester. We will refer to these groups as low, medium, and high interaction users. In the Fall dataset, the low interaction users averaged 8.6 utterances per student, and the high interaction users averaged 231.8 utterances per student. In the Spring, the low interaction averaged only 3.2 utterances per student, and the high interaction users averaged 284 utterances per student. 

To assess how interaction levels relate to student outcomes, we run multiple linear regression models. Identically to the previous section, we add a control independent variable to each model (the running average of prior assessment scores), use a dependent variable that is the normalized score of the assessment in the range $[0,1]$, and run this configuration for our baseline regressions. For group membership, we include three additional independent binary variables that indicate whether the student was in the low, medium, or high interaction group and conduct these regressions separately for the Fall and Spring semesters.

The baseline mean $R^2$ scores across all regressions are 0.1783 for the fall semester and 0.1682 for the spring semester. Adding the additional group features increases the mean $R^2$ to 0.2032 and 0.1858, respectively. This additional categorical variable yields a borderline significant improvement in explained variance for one assessment, Spring e1, $F = 4.08$, $p = 0.05$. On this assessment, the average scores for the low interaction, medium interaction, and high interaction users are 77.8\%, 81.0\%, and 88.1\%, respectively. Corroborating with our finding in the previous section, it appears that students with high usage, focused on conceptual questions on early assignments, tend to perform better on the first exam. While the grouping-level features are statistically noteworthy, the modest gains suggest that usage patterns have limited explanatory power for the overall course outcome.

We compare each group by plotting individual students' average assignment scores and average exam scores, as shown in Figure~\ref{fig:llm_nollm}. We observe the minimum average assignment score (92.53), and the minimum average exam scores (72.33) for the high interaction users across both semesters are higher than the minimum of the other two groups. Moreover, the variance in the average exam and assignment scores is smaller for high interaction users. For the remaining 90\% of users (low and medium interaction users), there was no substantial difference in minimum scores or variance between groups.  Our analysis demonstrates that although interaction level is not a dominant predictor of outcome, high engagement with the LLM is associated with reduced variability in course outcomes and higher minimum course outcomes. This pattern suggests that encouraging consistent and meaningful LLM usage, particularly among students who need additional tutoring and guidance, may help stabilize performance and raise lower-end outcomes, even if the overall effect on average scores is modest.

\subsection{Clustering Students by Behavior}
\begin{figure*}[ht!]
\centering
\begin{minipage}{\linewidth}
    \centering
    \includegraphics[width=0.85\linewidth]{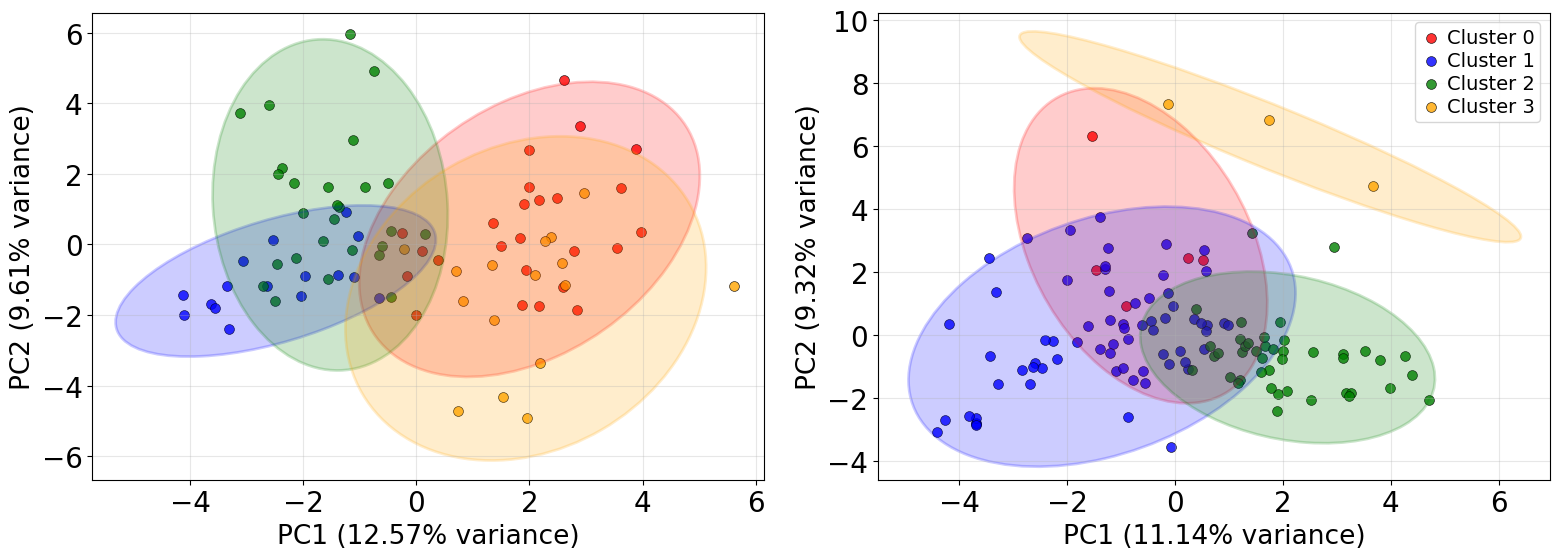}
\end{minipage}

\vspace{1em} % add a little space between top and bottom

\begin{minipage}{\linewidth}
    \centering
    \setlength{\tabcolsep}{3pt}
    \renewcommand{\arraystretch}{1.1}
    \begin{tabular}{l *{4}{S[table-format=2.1]} *{4}{S[table-format=2.1]}}
    % \toprule
    & \multicolumn{4}{c}{\textbf{Fall 2024}} & \multicolumn{4}{c}{\textbf{Spring 2025}} \\
    \cmidrule(lr){2-5} \cmidrule(lr){6-9}
    \textbf{Specific Dialogue Act} & {\textbf{C0}} & {\textbf{C1}} & {\textbf{C2}} & {\textbf{C3}}
                                   & {\textbf{C0}} & {\textbf{C1}} & {\textbf{C2}} & {\textbf{C3}} \\
    \midrule
    Conceptual Questions: Python Library           &  4.2 & 26.3 & 16.0 &  2.4 &  1.2 &  3.0 & 15.7 & 0.0 \\
    Writing Request: Write Code                    & 24.4 & 11.3 &  9.3 & 17.8 &  7.9 & 22.3 & 12.1 & 0.0 \\
    Writing Request: Write English                 &  8.6 &  1.7 &  4.0 & 16.8 &  5.8 & 13.0 &  5.1 & 0.0 \\
    Conceptual Questions: Programming Language     &  4.4 & 15.6 &  6.5 &  1.5 &  0.9 &  1.4 & 10.2 & 0.0 \\
    Writing Request: Code/Data Conversion          &  2.6 &  2.1 &  4.0 & 14.5 &  0.6 &  7.1 &  3.4 & 0.0 \\
    Conceptual Questions: Computer Science         &  4.1 & 13.1 &  7.7 &  8.1 &  2.6 &  4.3 & 11.3 & 0.0 \\
    Contextual Questions: Code Explanation         &  3.7 &  4.0 &  9.4 &  2.7 &  4.3 &  4.6 &  6.3 & 3.0 \\
    Contextual Questions: Assignment Clarification &  5.9 &  2.7 &  8.8 &  3.2 &  4.1 & 10.3 &  6.3 & 3.0 \\
    Provide Context: Error Message                 &  8.0 &  3.2 &  2.9 &  1.8 &  8.5 &  5.0 &  3.5 & 0.0 \\
    Provide Context: Other                         &  7.4 &  2.3 &  3.9 &  3.7 & 23.9 &  4.8 &  3.1 & 0.0 \\
        \midrule
\textbf{Cluster Size}
& \multicolumn{1}{c}{25}
& \multicolumn{1}{c}{16}
& \multicolumn{1}{c}{27}
& \multicolumn{1}{c}{16}
& \multicolumn{1}{c}{5}
& \multicolumn{1}{c}{68}
& \multicolumn{1}{c}{43}
& \multicolumn{1}{c}{3} \\
    \bottomrule
    \end{tabular}
\end{minipage}

\caption{PCA of DA features for Fall (top-left) and Spring (top-right). The table (bottom) summarizes the cluster distributions and shows the percentage of specific dialogue acts for the top 10 most common specific DA across all clusters. We see the type of question and writing requests as consistent distinguishing features across clusters. See appendix for further cluster details.}
\label{fig:pca_plus_table}
\end{figure*}

To investigate whether distinct patterns of student behavior emerge, we apply unsupervised clustering to behavioral interaction features. We follow an approach similar to \cite{brender2024s}, forming feature vectors that normalize each student's DA utterance types by their total count of utterances. In total, there are 39 features, with the 8 broad DA features and 31 specific features. These vectors represent each student's distribution of DA labels over the entire semester. We use K-means clustering on these feature vectors to group students based on their behavioral profiles, applying the method separately to Fall and Spring semester data. The optimal number of clusters, $k = 4$, is determined using the elbow method and silhouette analysis.

We visualize these groupings with Principal Component Analysis (PCA) in Figure~\ref{fig:pca_plus_table}, reducing the behavioral feature space to two components that captured the largest sources of variation (Fall: PC1 = 12.6\%, PC2 = 9.6\%; Spring: PC1 = 11.1\%, PC2 = 9.3\%). PC1 contrasts technical question–oriented usage (conceptual questions, Python library, programming language queries) with writing-focused (writing code) and context-providing behaviors; PC2 contrasts off-topic behavior (chit chat, greeting) with writing (write English, write code) and conceptual questions. Overlaid cluster boundaries and labels in the plots highlight these behavior styles and their degree of separation.

We report the distribution of the most common DA for each cluster in the table at the bottom of Figure~\ref{fig:pca_plus_table}. While the exact makeup of each cluster varies across semesters, some general patterns emerge. For example, certain broad labels, such as \textit{writing request} and \textit{contextual question} labels, are prevalent across multiple clusters, while others (e.g., \textit{off-topic} or \textit{conceptual questions}) are concentrated in specific clusters. Notably, some clusters represent high engagement (e.g., students with over 100 total utterances), while others capture minimal or off-topic usage. These differences suggest a spectrum of interaction styles, from consistent academic engagement to sparse or misaligned LLM usage. We see two recurring patterns of cluster assignment: groups with frequent writing requests and groups with frequent question-asking requests.

\begin{figure*}[t]
  \centering
  % Left: image
  \begin{minipage}[c]{0.9\textwidth}
    \centering
    \includegraphics[width=\linewidth]{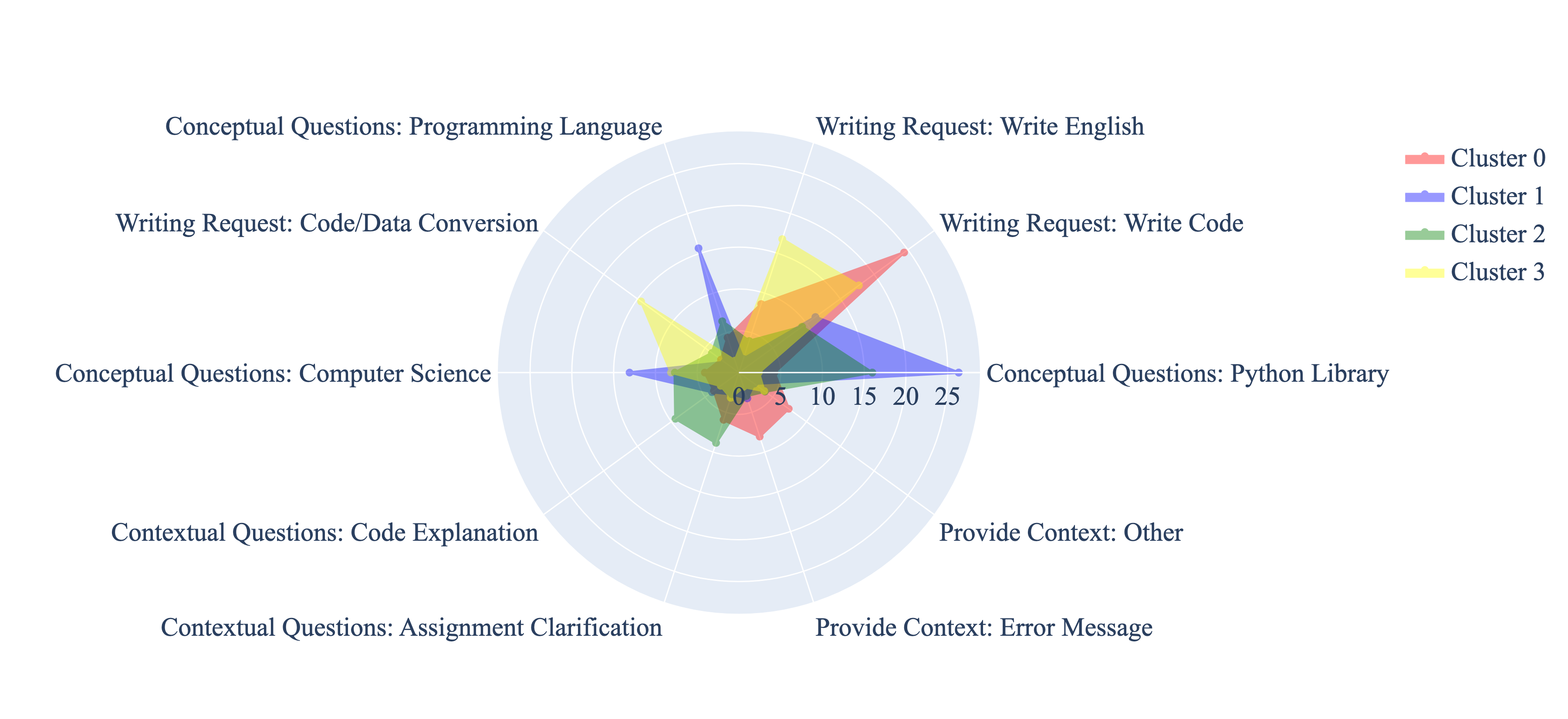}
  \end{minipage}\hfill
  % Right: table
  \begin{minipage}[c]{0.65\textwidth}
    \centering
    \small
    \renewcommand{\arraystretch}{1.1}
    \setlength{\tabcolsep}{4pt}
    \begin{tabular}{@{}r l r l l@{}}
      \toprule
      \# & Cluster          & n  & Exam              & Assign.         \\
      \midrule
      0  & Code Writers     & 25 & \(87.0 \pm 0.6\)  & \(95.6 \pm 2.5\) \\
      1  & Coding Question  & 16 & \(89.0 \pm 7.3\)  & \(96.1 \pm 2.1\) \\
      2  & General Question & 27 & \(86.5 \pm 8.0\)  & \(94.4 \pm 6.3\) \\
      3  & Report Writers   & 16 & \(83.8 \pm 11.7\) & \(93.7 \pm 4.4\) \\
      \bottomrule
    \end{tabular}
  \end{minipage}

  \caption{Comparison of dialogue act PCA clusters from Fall 2024. The radar plot (top) shows the top 10 most prominent specific dialog acts with values indicating the mean percentage of that act type within each cluster. The table (bottom) reports the number of students in each cluster and their mean aggregate course scores. Exam scores of cluster 3 are lower on average than other clusters.}
  \label{fig:clusters_radar_table}
\end{figure*}

We visualize the most predominant DAs and student grades for the fall semester PCA clusters in Figure~\ref{fig:clusters_radar_table}. We observe further patterns in examining the behavior patterns and course outcomes in these clusters. For the writing-request clusters, students in cluster 0 tend to ask the chatbot to write code, while students in cluster 3 display a mixture of code writing and English writing requests. This observation suggests that students in cluster 3 lean on the LLM not only for help in implementing the requested program but also in synthesizing the results and writing assignment reports. For the question-asking clusters, students in cluster 1 tend to ask conceptual questions, mostly on Python-related questions and computer science topics, while students in cluster 2 ask a mixture of conceptual questions and contextual questions of a much wider spread of topics. This observation suggests that students in cluster 1 use the LLM for understanding the tools needed to complete the assignments, while students in the latter category use the LLM as a more general-purpose tutoring system. In examining the average assignment and exam scores between these clusters, we see that all clusters have comparable average assignment scores. However, coding-question groups trend towards a higher average score on the exams $89\pm7.3$ while report-writers tend to have a marginally lower average score $83.8\pm11.7$. These observations suggest that students who ask targeted questions of the LLM for specific programming and coding concepts while working on assignments tend to enter exams with a better understanding of the material than students who request extensive writing assistance, which may circumvent the learning objectives of the assignments.

To further explore these trends in assessment scores, we again perform a regression analysis. We use the same baseline model, a single independent variable, running average of prior assessment scores, to dependent variable of assessment grades. Similar to the usage interaction analysis, for cluster membership, we include four additional binary variables, each associated with a group to assess its additional explanatory power. We find that adding the cluster variable improved the predictive performance marginally. For the Fall semester, the mean $R^2$ increased from 0.1783 to 0.2152; for Spring, it improved from 0.1682 to 0.1844. We conduct F-tests for each regression and find no indications of statistically significant improvement in explanatory power, with the highest resulting F-score of F=2.67, p=0.06. These results demonstrate that while this behavioral clustering does provide an improvement, similar to the usage grouping, we cannot definitively conclude that this clustering strategy is statistically meaningful in terms of understanding course outcomes. However, while the clusters are not definitive predictors, they do offer interpretable groupings of student behavior that correlate with distinct behavior styles ranging from frequent, academically aligned interactions to sparse or off-topic usage. 
% This diversity highlights that engagement is multidimensional, varying not only in frequency but also in the type of behavior. 
While our clustering analysis reveals some behavior types that are associated with lower exam performance, further work is needed to achieve consistent and reliable detection of behaviors that are misaligned with learning objectives.
% These clusters provide quantitative insights into patterns of student LLM usage, which may prove helpful in educating students to use LLM tools in a learning-positive way.

\section{Conclusions, Limitations, Future Work}

In this work, we present the StudyChat dataset, a collection of over 16,000  interactions between students and ChatGPT in a university-level artificial intelligence course. We systematically create and apply a dialogue act labeling schema applied to this dataset via LLM labeling, validate this labeling with human agreement, and conduct regression and clustering analysis to analyze trends of student behavior as they relate to course outcomes.

While our analysis offers new insights into how students interact with LLMs, the takeaways from our work have limitations. First, our dataset was drawn from a single upper-division AI course, which limits how well the findings generalize. Second, participants in this study were aware that their conversations were recorded and thus their behaviors may have been altered and subject to the Hawthorne Effect \cite{landsberger1958hawthorne}.

Our findings from this exploration and the resulting dataset open many avenues for future work. First, since the StudyChats dataset also includes student assignment submissions, we can further compare these submissions against student utterances when conversing with the LLM, which may reveal further behavioral insights and trends. Second, our initial labeling schema and labeling methodology could be refined further to better cover overlapping labels and missing dialogue acts with more accurate and fine-grained methods of classifying student utterances. Finally, these insights can be taken back into a classroom setting where we could create LLM-based tutoring tools that detect and discourage behaviors associated with negative course outcomes.

\begin{acks}
We would like to thank all the students of this course for their participation and engagement in this project. We are also grateful to the entire course staff for their hard work and dedication to teaching.
We extend a special thanks to Aditya Singh for his contributions to the design and implementation of many course assignments, as well as the insightful discussions he shared with the authors regarding this work.
The authors are partially supported by the NSF under grants 2237676 and 2418657.
\end{acks}

\bibliographystyle{ACM-Reference-Format}
\bibliography{bibliography}

\clearpage
\appendix
\onecolumn
\section{Appendix}

\begin{table}[h]
\centering
\renewcommand{\arraystretch}{1.2}
\setlength{\tabcolsep}{6pt}
\caption{Distribution of dialogue act labels in StudyChat.}
\label{tab:dialogue_act_distribution}
\begin{tabular}{l r | l r | l r}
\toprule
\multicolumn{2}{c|}{\textbf{Conceptual Questions}}  & \multicolumn{2}{c|}{\textbf{Contextual Questions}}  & \multicolumn{2}{c}{\textbf{Writing}} \\
\midrule
Python Library & 1788  & Assignment Clarification & 1076  & Write Code      & 2534  \\
Programming Language  & 1271   & Code Explanation        & 950  & Write English    & 1340  \\
Computer Science     & 1178  & Interpret Output        & 513   & Conversion       & 347  \\
Programming Tools    & 484   & Other                   & 399 & Summarize        & 52   \\
Mathematics          & 155  &                         &      & Other            & 63  \\
Other Concept        & 323  &                         &      &                  &   \\
\textbf{Total}       & \textbf{5199} & \textbf{Total} & \textbf{2938} & \textbf{Total}  & \textbf{4336
} \\
\midrule
\multicolumn{2}{c|}{\textbf{Verification}} & \multicolumn{2}{c|}{\textbf{Context}}  & \multicolumn{2}{c}{\textbf{Editing }} \\
\midrule
Verify Code         & 420  & Assignment Information   & 434  & Edit Code        & 213  \\
Verify Report       & 100   & Error Message           & 979  & Edit English     & 197  \\
Verify Output       & 77   & Code                    & 696  & Other            &  8    \\
Other               & 50   & Other                   & 919  &                  &  \\
\textbf{Total}      & \textbf{647}  & \textbf{Total}  & \textbf{3028} & \textbf{Total}    &  \textbf{418}    \\
\midrule
\multicolumn{2}{c|}{\textbf{Misc}} & \multicolumn{2}{c|}{\textbf{Off Topic}} & \multicolumn{2}{c}{} \\
\midrule
Other & 30 & Chit-Chat & 102 &  \\
\textbf{Total} & \textbf{30} & Gratitude & 35 &  \\
 & & Greeting & 84 &  \\
 & & Other & 29 &  \\
 & & \textbf{Total} & \textbf{250} &  \\
\bottomrule
\end{tabular}
\end{table}

\begin{table*}[ht!]
\centering
\renewcommand{\arraystretch}{1.15}
\setlength{\tabcolsep}{6pt}
\caption{Top 10 Dialogue Acts per Cluster -- F24 Semester}
\label{tab:f24_clusters}
\begin{tabular}{lllr}
\toprule
\textbf{Cluster} & \textbf{Users} & \textbf{Dialogue Act} & \textbf{\% Occurrence} \\
\midrule
\multirow{10}{*}{Cluster 0} & \multirow{10}{*}{25} 
 & Broad Writing Request & 36.66 \\
 & & Broad Provide Context & 24.73 \\
 & & Specific Writing Request Write Code & 24.44 \\
 & & Broad Conceptual Questions & 17.43 \\
 & & Broad Contextual Questions & 14.98 \\
 & & Specific Writing Request Write English & 8.63 \\
 & & Specific Provide Context Error Message & 8.05 \\
 & & Specific Provide Context Other & 7.37 \\
 & & Specific Contextual Questions Assignment Clarification & 5.94 \\
 & & Specific Provide Context Code & 4.82 \\
\midrule
\multirow{10}{*}{Cluster 1} & \multirow{10}{*}{16} 
 & Broad Conceptual Questions & 63.02 \\
 & & Specific Conceptual Questions Python Library & 26.31 \\
 & & Specific Conceptual Questions Programming Language & 15.64 \\
 & & Broad Writing Request & 15.32 \\
 & & Specific Conceptual Questions Computer Science & 13.09 \\
 & & Specific Writing Request Write Code & 11.32 \\
 & & Broad Contextual Questions & 10.84 \\
 & & Broad Provide Context & 8.24 \\
 & & Specific Conceptual Questions Programming Tools & 4.04 \\
 & & Specific Contextual Questions Code Explanation & 3.98 \\
\midrule
\multirow{10}{*}{Cluster 2} & \multirow{10}{*}{27} 
 & Broad Conceptual Questions & 36.20 \\
 & & Broad Contextual Questions & 25.88 \\
 & & Broad Writing Request & 17.89 \\
 & & Specific Conceptual Questions Python Library & 15.97 \\
 & & Broad Provide Context & 9.85 \\
 & & Specific Contextual Questions Code Explanation & 9.39 \\
 & & Specific Writing Request Write Code & 9.35 \\
 & & Specific Contextual Questions Assignment Clarification & 8.85 \\
 & & Specific Conceptual Questions Computer Science & 7.67 \\
 & & Specific Conceptual Questions Programming Language & 6.46 \\
\midrule
\multirow{10}{*}{Cluster 3} & \multirow{10}{*}{16} 
 & Broad Writing Request & 50.25 \\
 & & Specific Writing Request Write Code & 17.75 \\
 & & Broad Conceptual Questions & 17.18 \\
 & & Specific Writing Request Write English & 16.79 \\
 & & Specific Writing Request Code/Data Conversion & 14.47 \\
 & & Broad Provide Context & 10.72 \\
 & & Broad Contextual Questions & 8.14 \\
 & & Specific Conceptual Questions Computer Science & 8.13 \\
 & & Broad Verification & 6.61 \\
 & & Specific Verification Verify Code & 5.42 \\
\bottomrule
\end{tabular}
\end{table*}

\begin{table*}[ht!]
\centering
\renewcommand{\arraystretch}{1.15}
\setlength{\tabcolsep}{6pt}
\caption{Top 10 Dialogue Acts per Cluster -- S25 Semester}
\label{tab:s25_clusters}
\begin{tabular}{lllr}
\toprule
\textbf{Cluster} & \textbf{Users} & \textbf{Dialogue Act} & \textbf{\% Occurrence} \\
\midrule
\multirow{10}{*}{Cluster 0} & \multirow{10}{*}{5} 
 & Broad Provide Context & 46.85 \\
 & & Specific Provide Context Other & 23.95 \\
 & & Broad Writing Request & 14.63 \\
 & & Broad Conceptual Questions & 13.12 \\
 & & Broad Off Topic & 10.88 \\
 & & Specific Off Topic Greeting & 10.63 \\
 & & Specific Provide Context Code & 10.11 \\
 & & Broad Contextual Questions & 9.95 \\
 & & Specific Provide Context Error Message & 8.49 \\
 & & Specific Writing Request Write Code & 7.95 \\
\midrule
\multirow{10}{*}{Cluster 1} & \multirow{10}{*}{68} 
 & Broad Writing Request & 44.63 \\
 & & Specific Writing Request Write Code & 22.28 \\
 & & Broad Contextual Questions & 18.46 \\
 & & Broad Provide Context & 17.04 \\
 & & Specific Writing Request Write English & 12.99 \\
 & & Broad Conceptual Questions & 10.75 \\
 & & Specific Contextual Questions Assignment Clarification & 10.30 \\
 & & Specific Writing Request Code/Data Conversion & 7.13 \\
 & & Specific Provide Context Error Message & 5.00 \\
 & & Specific Provide Context Other & 4.76 \\
\midrule
\multirow{10}{*}{Cluster 2} & \multirow{10}{*}{43} 
 & Broad Conceptual Questions & 43.55 \\
 & & Broad Writing Request & 21.25 \\
 & & Broad Contextual Questions & 17.97 \\
 & & Specific Conceptual Questions Python Library & 15.67 \\
 & & Specific Writing Request Write Code & 12.11 \\
 & & Specific Conceptual Questions Computer Science & 11.29 \\
 & & Broad Provide Context & 10.44 \\
 & & Specific Conceptual Questions Programming Language & 10.19 \\
 & & Specific Contextual Questions Assignment Clarification & 6.31 \\
 & & Specific Contextual Questions Code Explanation & 6.27 \\
\midrule
\multirow{10}{*}{Cluster 3} & \multirow{10}{*}{3} 
 & Broad Off Topic & 41.82 \\
 & & Specific Off Topic Chit-Chat & 29.09 \\
 & & Broad Contextual Questions & 26.06 \\
 & & Broad Conceptual Questions & 23.03 \\
 & & Specific Conceptual Questions Other Concept & 20.00 \\
 & & Specific Contextual Questions Other & 20.00 \\
 & & Specific Off Topic Greeting & 12.73 \\
 & & Broad Provide Context & 3.03 \\
 & & Broad Verification & 3.03 \\
 & & Broad Writing Request & 3.03 \\
\bottomrule
\end{tabular}
\end{table*}

\begin{figure}[h]
    \centering
    \includegraphics[width=\linewidth]{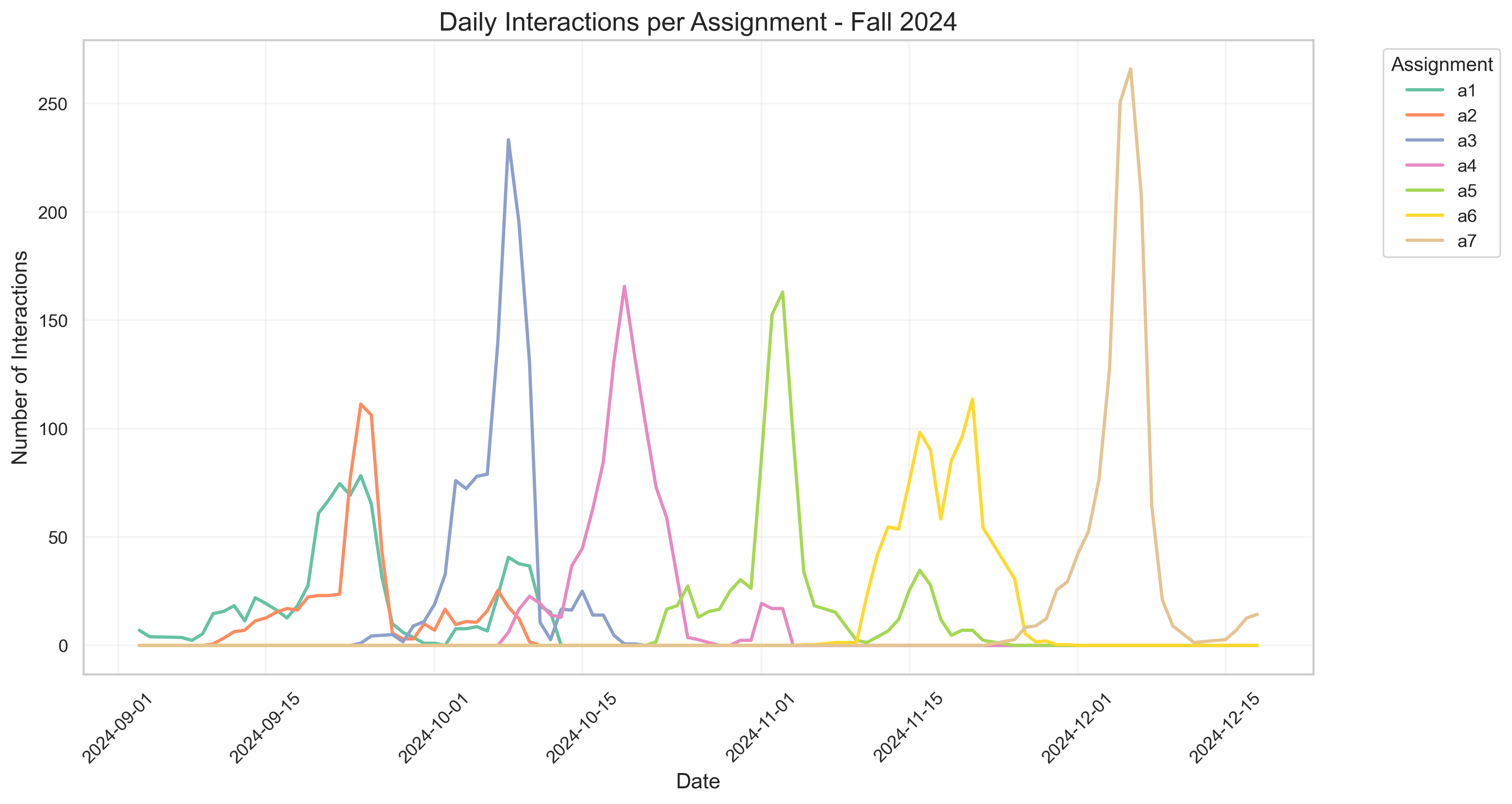}
    \caption{Usage over time organized by assignments for the Fall 2024 semester}
    \label{fig:daily_usage_fall}

    \vspace{1em} % optional space between figures

    \includegraphics[width=\linewidth]{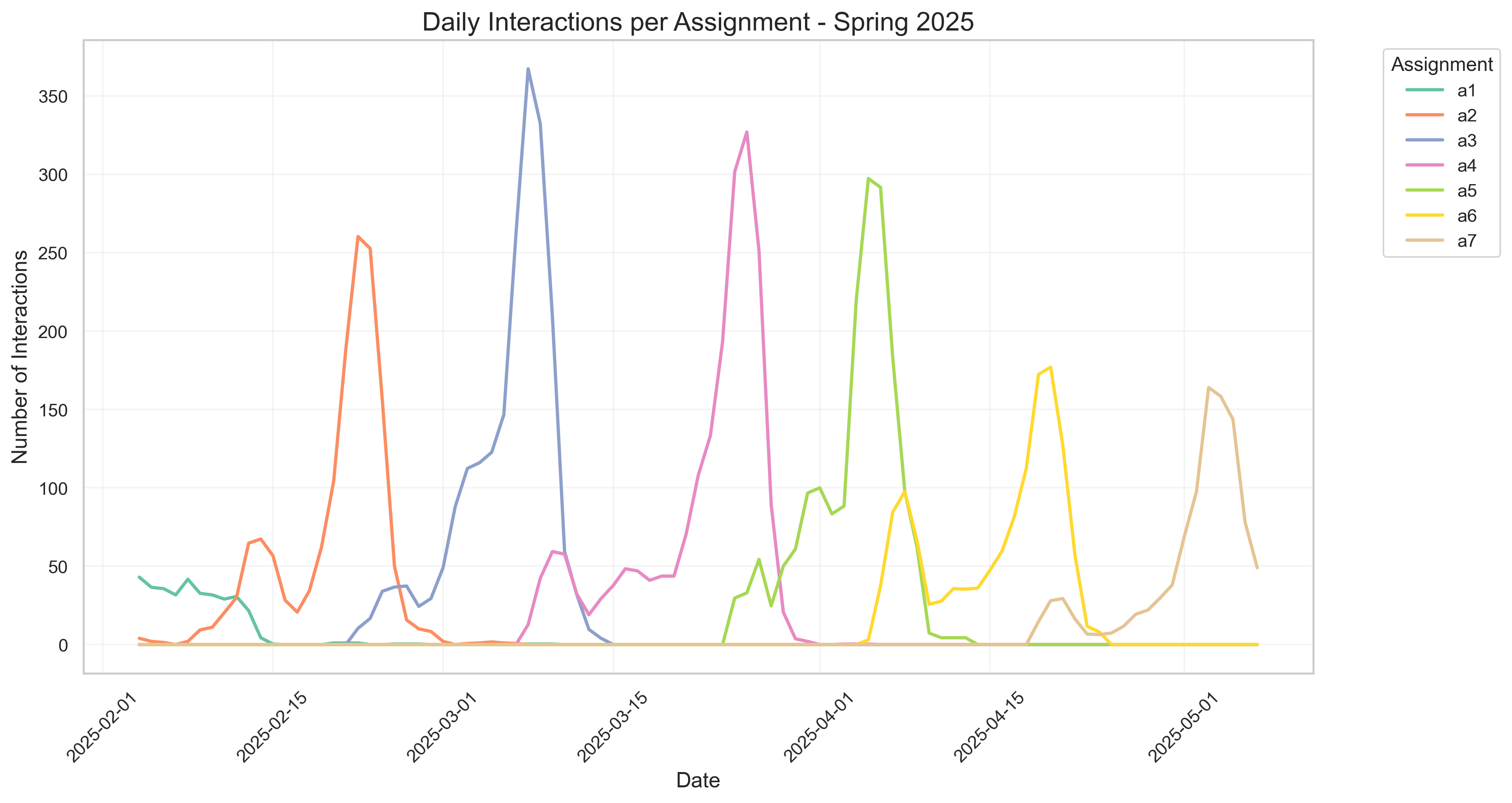}
    \caption{Usage over time organized by assignments for the Spring 2025 semester}
    \label{fig:daily_usage_spring}
\end{figure}

\begin{figure}[h] % figure* no longer needed in one-column mode
    \centering
    \includegraphics[width=0.75\linewidth]{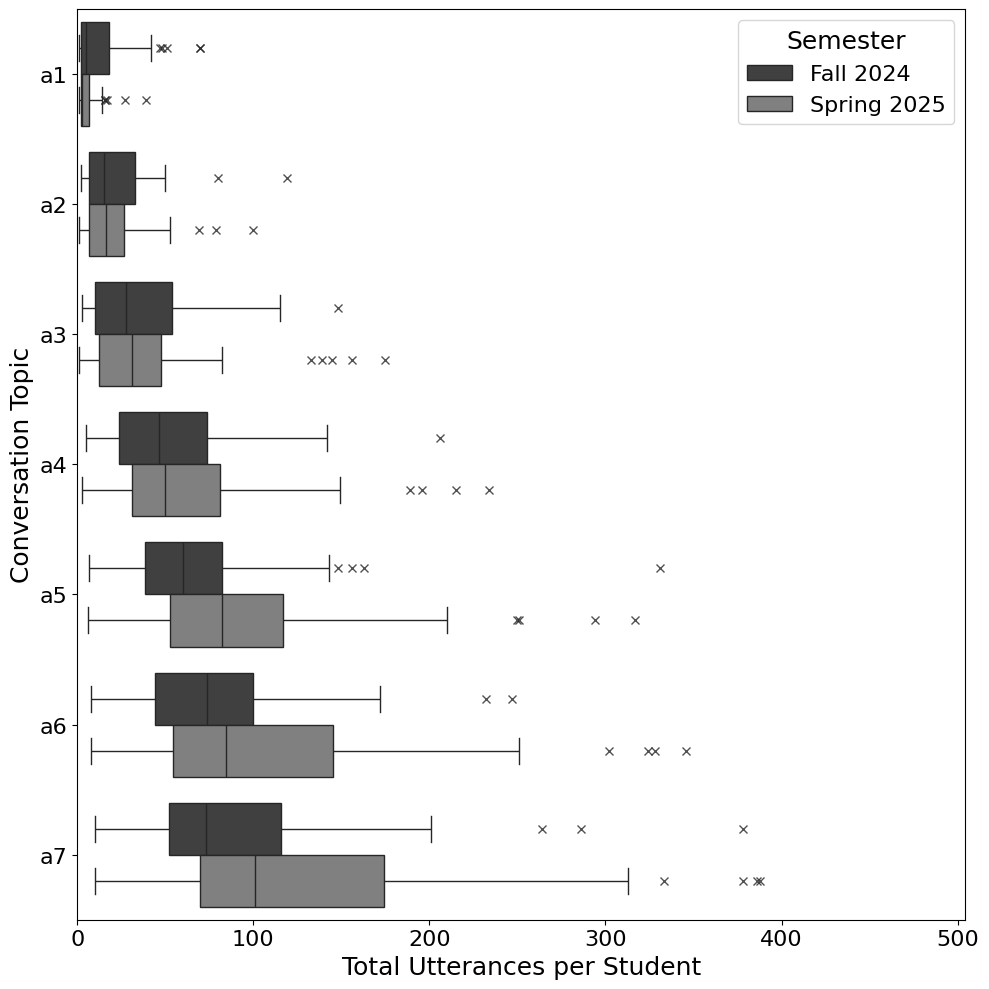}
    \caption{Box plot showing the distribution of total utterances per student for each conversation topic where outliers are shown with cross markers.}
    \label{fig:utterance_violin}
\end{figure}

\end{document}